\documentclass[5p,twocolumn,nopreprintline]{elsarticle}

\usepackage{lineno,hyperref}
\modulolinenumbers[5]

\usepackage{amsmath}
\usepackage{amsfonts}
\usepackage{graphicx}
\usepackage{float}
\usepackage{booktabs}
\usepackage{multirow}
\usepackage[table,xcdraw]{xcolor}
\usepackage{overpic}
\usepackage{lineno}  
\usepackage{algorithm}
\usepackage{algorithmic}
\usepackage{multicol}

\newtheorem{proposition}{Proposition}

\newtheorem{remark}{Remark}
\newtheorem{corollary}{Corollary}

\biboptions{authoryear}

\begin{document}

\begin{frontmatter}

\title{Self-supervised Multi-task Learning Framework for Safety and Health-Oriented Connected Driving Environment Perception using Onboard Camera}



\author[address1,address3]{Shaocheng JIA}

\author[address1,address2]{Wei YAO\corref{mycorrespondingauthor}}
\cortext[mycorrespondingauthor]{Corresponding author.}

\address[address1]{Department of Land Surveying and Geo-Informatics, The Hong Kong Polytechnic University, Hung Hom, Hong Kong}
\address[address2]{The Hong Kong Polytechnic University Shenzhen Research Institute, Shenzhen, China}
\address[address3]{Department of Civil Engineering, The University of Hong Kong, Hong Kong}


\begin{abstract}
Cutting-edge connected vehicle (CV) technologies have drawn much attention in recent years. The real-time traffic data captured by a CV can be shared with other CVs and data centers so as to open new possibilities for solving diverse transportation problems. The trajectory data of CVs have been well-studied and widely used. However, image data captured by onboard cameras in a connected environment, as being a kind of basic data source, are not sufficiently investigated, especially for safety and health-oriented visual perception. In this paper, a bidirectional process of image synthesis and decomposition (BPISD) approach is proposed, and thus a novel self-supervised multi-task learning framework, to simultaneously estimate depth map, atmospheric visibility, airlight, and PM$_{2.5}$ mass concentration, in which depth map and visibility are considered highly associated with traffic safety, while airlight and PM$_{2.5}$ mass concentration are directly correlated with human health. Both the training and testing phases of the proposed system solely require a single image as input. Due to the innovative training pipeline, the depth estimation network can automatically manage various levels of visibility conditions and overcome diverse inherent problems in current image-synthesis-based self-supervised depth estimation, thereby generating high-quality depth maps even in low-visibility situations and further benefiting accurate estimations of visibility, airlight, and PM$_{2.5}$ mass concentration. Extensive experiments on the original and synthesized data from the KITTI dataset and real-world data collected in Beijing demonstrate that the proposed method can (1) achieve performance competitive in self-supervised depth estimation as compared with other state-of-the-art methods when taking clear images as input; (2) predict vivid depth map for images contaminated by various levels of haze when the network trained with previous framework fails; and (3) accurately estimate visibility, airlight, and PM$_{2.5}$ mass concentrations. Beneficial applications can be developed based on the presented work to improve traffic safety, air quality, and public health.

\end{abstract}


\end{frontmatter}


\section{Introduction} 
In the past decade, communication technologies have undergone substantial development. In transportation systems, connected vehicle (CV) technology enables information exchanges between different system components. The running connected vehicles can continuously collect data of interest and send them back to the data center for further analysis. Taking CVs as mobile sensors opens new possibilities to solve transportation problems using the shared trajectory data \citep{jia2023uncertainty}. It provides us insights to conduct driving environment perception tasks in a connected environment. Specifically, the image data captured by the onboard cameras installed on the CVs can also be shared with other CVs and stored in the data centers. As the road network generally matches the whole city, CVs, therefore, are distributed in large-scale urban areas due to regular traffic demands. Together with the fact that CV flows are continuous in time, the quantity estimations can then be spatiotemporally continuous. This may promote paradigm shifts in driving environment perception.

The following four types of information are considered important for driving environment perception. First, depth estimation is crucial for 3D scene understanding and the driver's decision-making. While current depth estimation methods require either truthful labels for supervision or image sequence for self-supervised training. The former is expensive. The latter may fail when the static scene assumption is violated in a driving environment. Second, visibility is highly associated with traffic safety. Moreover, PM$_{2.5}$ mass concentration and airlight are related to air quality, and thus human health. However, visibility, PM$_{2.5}$ mass concentration, and airlight are often measured with a professional instrument outfitted on the roads. Due to the high capital costs, it is impossible to obtain dense estimates in a large-scale city. Moreover, substantial differences in these quantities can exist across various urban areas, raising from diverse land uses, special geographical properties, and complex weather conditions. Therefore, spatiotemporally continuous estimates are expected for achieving more accurate driving environment perception. The dynamic CV signals offer a unique channel for solving these problems.

Taking the image data collected by CVs as input, a self-supervised co-training framework is proposed to simultaneously estimate depth map, airlight, clear image, and visibility using four convolutional neural networks (CNNs). The forward inference process is an image decomposition phase. The estimated results are coupled with Koschmieder’s law to reconstruct the input image. This reconstruction process is an image synthesis phase. The difference between the input image and the reconstructed image is used to train the system. It is noted that after co-training each sub-network can be used separately. Upon the estimated visibility, a statistical model is further deployed to map visibility with PM$_{2.5}$ mass concentration, where low visibility is assumed to be solely caused by PM$_{2.5}$. The estimated PM$_{2.5}$ mass concentrations are projected back to the physical world in terms of the location information for precise air quality monitoring. As traffic is highly dynamic, the estimated local PM$_{2.5}$ mass concentration can be continuously updated. The expectation of multiple estimates for the same area can be taken as the final outcome. Visibility, depth map, and airlight, can also be projected onto the physical map and used in driving environment reconstruction or air component analysis. 

Such a bidirectional process of image synthesis and decomposition (BPISD) is radically different from the previous training pipeline wherein only image synthesis is used. Furthermore, the reference images used for reconstruction in the previous training pipeline have certain time and space shifts as compared with the target image, i.e., the image sequence is used. This introduces issues of occlusion, moving objects, lighting changes, etc. The presented work solely requires a single image as input. The corresponding clear image (i.e., dehazed image in this study) will be the reference image for reconstruction. All the above-mentioned issues are automatically resolved.

To validate the proposed system, a large number of driving-view images with various visibilities caused by different degrees of PM$_{2.5}$ mass concentrations are needed for training the deep neural networks. Such a dataset, however, is unavailable to the best of our knowledge. To address this issue, a novel synthetic method based on Koschmieder’s law has been proposed and applied to Zhou et al.’ split \citep{zhou2017unsupervised} in self-supervised depth estimation on the KITTI dataset \citep{geiger2013vision}. Eigen’s split \citep{zhou2017unsupervised} of the KITTI dataset was used for evaluating the performance of depth estimation. The trained visibility estimation model was directly applied to the real-world data collected in Beijing without any refinement. Adopting simple polynomial fitting, the estimated visibilities can be readily transformed to PM$_{2.5}$ mass concentrations. Comparisons show that the proposed method achieves competitive and robust performance in self-supervised depth estimation under various visibility conditions. For estimation of visibility and PM$_{2.5}$ mass concentrations, it was found that the mean absolute percentage errors (MAPE) were respectively confined within $5\%$ and $8\%$ across various levels of relative humidity with the order of polynomial fitting beyond 6. These promising performances clearly demonstrate the effectiveness of the proposed method. It is noted that the said approach does not require any additional devices or change the vehicle configurations, being a non-intrusive method. As such, it has great potential to be implemented for monitoring real-time particle conditions and promoting paradigm shifts on many applications, e.g., starting and embracing health-aware navigation and travel planning.

In short, the contributions of this paper are fourfold:
\begin{itemize}
\item  A CV-based framework is proposed for estimating spatiotemporally continuous visibility, airlight, and PM$_{2.5}$ mass concentration in large-scale cities and possibly establishing dynamical 3D driving maps by means of accurate depth maps irrespective of diverse visibility conditions. 

\item A novel bidirectional process of image synthesis and decomposition (BPISD) paradigm is proposed, and thus a unified self-supervised multi-task learning framework, to simultaneously estimate depth map, visibility, airlight, and PM$_{2.5}$ mass concentration.

\item Leveraging Koschmieder’s law, a haze image synthesis model is introduced, which can conveniently generate various contaminated images for diverse use, e.g., training dehazing networks.

\item Applying the proposed methods to the KITTI dataset and the real-world Beijing data affords excellent performance in all four subtasks. In particular, the proposed depth estimation method can achieve performance competitive compared to state-of-the-art methods and significantly outperform the method trained with the traditional self-supervised pipeline in the case of low-visibility situations.

\end{itemize}

The remainder of this paper is organized as follows. Section $2$ reviews  related works. Section $3$ defines the problem. Section $4$ presents the proposed method. Section $5$ reports experimental results. Section $6$ concludes the paper.

\section{Related work}

\subsection{Estimation of depth map and visibility}
Depth map offers important three-dimensional (3D) information for the given image, and thus in conjunction with LiDAR point clouds \citep{WANG2022237} is widely used in 3D reconstruction, scene understanding, and autonomous driving. However, it is nontrivial to accurately estimate the depth map from a single image, as monocular depth estimation is an inherently ill-posed problem. Prior to the prosperity of deep learning, hand-crafted features need to be extracted from the input raw images. Then, regression or classification process is used to estimate the depth map \citep{saxena20083, baig2016coupled, choi2015depth, furukawa2017depth, zoran2015learning}. This type of method heavily relies on the feature design. It is challenging to devise abstract and comprehensive features manually. 

Deep learning, especially convolutional neural networks (CNN), can automatically extract abstract and deep features from the input images\citep{POLEWSKI2021297}, and thus provide new channels to conduct monocular depth estimation. Given the ground truth depth map, the end-to-end networks are trained by minimizing the difference between the estimated and truthful depth maps. Tremendous work has been conducted focusing on exploring various CNN structures \citep{chen2016single,eigen2014depth,eigen2015predicting,laina2016deeper,li2017two} and capturing global information of the images \cite{cao2017estimating,eigen2015predicting,li2015depth,liu2015deep,mousavian2016joint,xu2017multi,xu2018structured, almalioglu2019ganvo, cs2018depthnet, grigorev2017depth, mancini2017toward, tananaev2018temporally, wang2019recurrent, mypaper2020}. However, such supervised depth map methods require a large number of truthful depth maps. This hinders the models' universal application. 

Self-supervised depth estimation does not require any labeled data for training the networks. The previous training pipeline requires continuous image sequences as input. By defining target and reference images, the difference between the reconstructed and original target images is computed for training the whole system, in which the reconstruction of target images is based on reference images, the depth map of the target image, and the ego-motion between the target image and reference images. This pipeline has attracted many researchers \citep{zhou2017unsupervised, chen2019towards, garg2016unsupervised, ranjan2019competitive, yin2018geonet, zhan2018unsupervised, zhou2019unsupervised, zhou2017unsupervised, godard2017unsupervised, kuznietsov2017semi, almalioglu2019ganvo, feng2019sganvo, guizilini20203d, mypaper2021, mypaper2, jia2023joint}.
However, the underlying assumption is that all objects in the scene are static. This assumption is often violated in a driving environment. Moreover, the common occurrence of occlusion and disocclusion in the course of vehicle moving also brings difficulty in finding pixel correspondence. Although some works are devoted to mitigating such issues \citep{godard2019digging, casser2019depth,klingner2020self,shu2020feature}, the problems can hardly be completely resolved.

Different from the previous training pipeline used in depth estimation, this work proposes a novel self-supervised multi-task learning framework based on a bidirectional process of image synthesis and decomposition (BPISD), which solely requires a single image as input for training the whole system. Thus, the above-described issues are automatically resolved, thereby possibly achieving better performance. A single image input lets the system training be more flexible as well. Multi-task learning is considered more cost-effective in driving environment perception.

In visibility estimation, three types of methods are often adopted, i.e., traditional methods, statistical methods, and deep neural network (DNN)-based methods. Traditionally, visibility is measured by either manual observation or professional instruments. For the former, the estimate could be varied from observer to observer. For the latter, professional instrument is generally expensive and is impossible to be densely installed in large-scale cities \citep{chaabani2017neural, pomerleau1997visibility}. Statistical methods estimates visibility by definition or modeling the relationship between the collected data and visibility \citep{dietz2019forecasting, cheng2018expressway}, which generally need to perform geographic calibration and thus are difficult for universal application. DNN-based methods can establish an end-to-end model for conventionally estimating visibility. While a large number of labeled data are required for training the networks \citep{palvanov2018dhcnn,you2022dmrvisnet}. 

The presented visibility estimation method differs from the previous works in two aspects: 1) the proposed method does not need any labeled data as supervision for training the networks, and 2) the spatiotemporally continuous visibility across a large-scale city can be estimated via the active CVs. This provides a unique opportunity for developing many real-time weather-related applications.

\subsection{Estimation of PM$_{2.5}$ mass concentration and airlight}

PM$_{2.5}$ refers to particulate matters (PM) that own aerodynamic diameters of no more than $2.5$ $\mu m$. Heterogeneous chemical compositions of PM$_{2.5}$ significantly impact aerosol light extinction including aerosol absorption and scattering \citep{xu2020current, cao2012impacts, tao2019impact}, and thus degenerate visibility \citep{renhe2014meteorological, wang2009clear, watson2002visibility} and influence transportation. Most importantly, PM$_{2.5}$ can be readily breathed into the lungs and further penetrate deep into the brain from the blood streams, causing serious health problems, e.g., cardiovascular disease, respiratory disease, and premature death \citep{us2016health}. With the rapid industrial development in many countries, such as China, India, and Nepal, quickly increasing energy consumption leads to deteriorating air quality in urban areas \citep{chan2008air, kan2012ambient}, especially in the level of PM$_{2.5}$ mass concentration. Regular haze weather draws special attention of both the general public and academia \citep{huang2020amplified}. Accurate and dynamical detection of PM$_{2.5}$ mass concentration becomes crucial for air quality monitoring and travel planning so as to protect public health. 

Various methods have been used in the estimation of PM$_{2.5}$ mass concentration. The commonest type of method is to devise and improve professional instruments based on various chemical principles \citep{chen2019aerosol,zhao2019recent,pandolfi2018european,malm2001estimates}. While the professional instrument can be costly and difficult for universal application. Considering the wide spatiotemporal distribution of PM$_{2.5}$, satellite-based remote sensing data were considered advantageous \citep{zheng2017analysis, chelani2019estimating, 10.1145/3429309.3429328, van2006estimating, sun2019deep}. However, the satellite data may not be economically and timely available. Moreover, several empirical models have also been proposed to estimate the PM$_{2.5}$ mass concentration from the atmospheric visibility \citep{wang2006quantitative, ji2020estimation}. Nevertheless, the atmospheric visibility still needs to be measured using professional instruments; this limits the models’ intensive use in dense estimations. Due to sophisticated meteorological changes and geographical differences between various urban areas, PM$_{2.5}$ mass concentrations exhibit high spatial-temporal variations in large-scale cities. The above-described instrument-based methods are hard to capture such variations in PM$_{2.5}$ mass concentration in light of the limited and fixed detector installations. Taking CVs as mobile "sensors", the proposed method can dynamically and densely estimate PM$_{2.5}$ mass concentrations for a whole city.

In airlight estimation, there are two types of methods in general: prior-based methods and learning-based methods. For the former, different priors have been studied, e.g., the brightest pixel prior \citep{he2010single, fattal2008single, tan2008visibility}, color constancy prior \citep{gautam2020improved}, color attenuation prior \citep{zhu2015fast}, statistical priors \citep{berman2016non, bahat2016blind, fattal2014dehazing}, etc. While various priors may have distinct application conditions; it is challenging for common use. The latter is to take the raw image as input and then directly estimate airlight through DNN \citep{cai2016dehazenet, zhang2018densely, wang2019airlight}. The presented method is also a learning-based method but, it does not require any labeled data and is simultaneously estimated along with the other three tasks. In this paper, airlight is considered associated with chemical components in the air, so that it can be used in air quality analysis for protecting human health.

\begin{figure*}[htp]
    \centering
    \includegraphics[width=\textwidth]{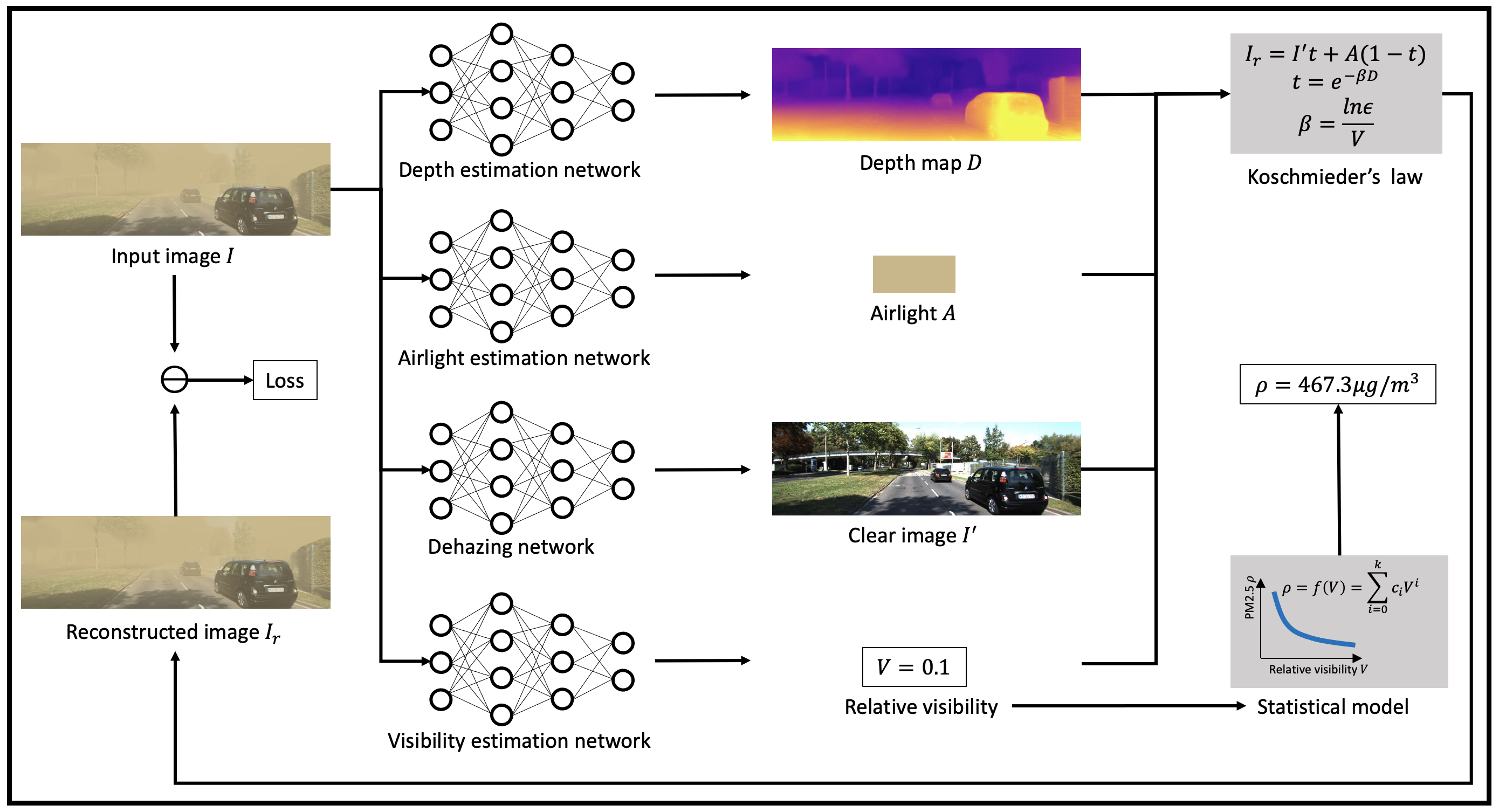}
    \caption{Architecture of the proposed framework.}
    \label{overview}
\end{figure*}

\section{Problem statement}
Given an image captured by the onboard camera, the goal is to simultaneously estimate the scene's depth map, visibility, airlight, and PM$_{2.5}$ mass concentration. Previous methods need to separately train the networks for estimating these quantities. For depth estimation, the image sequences are required for conducting self-supervised training. For the estimation of visibility, airlight, and PM$_{2.5}$ mass concentration, truthful labels are generally used for supervised training. One target of this study is to simultaneously estimate these quantities by devising a unified, self-supervised, and single-image training framework. 

Traditional methods for measuring visibility, airlight, and PM$_{2.5}$ mass concentration are based on fixed meteorological stations, which are sparsely distributed in the city, and thus are difficult to capture the spatial variations. Another goal of this study is to find a solution for densely estimating visibility, airlight, and PM$_{2.5}$ mass concentration in a megacity, thereby providing more precise perception information.

\section{Method}

This section introduces the detailed method. The overall architecture, hazy image synthesis model, self-supervised multi-task learning framework, and polynomial correlation model for PM$_{2.5}$ mass concentration estimation are presented, respectively.

\subsection{Architecture overview}
The whole system is trained in a self-supervised manner, meaning that no labeled data is required. Moreover, the system only takes a single image as input instead of an image sequence used in the previous self-supervised training framework. As shown in Figure \ref{overview}, the input image is passed to various sub-networks for estimating diverse information, including depth map, airlight, dehazed image, and visibility. This process is to conduct image decomposition. The estimates are then coupled with Koschmieder's law for reconstructing the input image; it is apparently an image synthesis process. The difference between the reconstructed image and the original input image is used to train the system. Upon the completion of visibility estimation, a statistical model is further utilized to map visibility to PM$_{2.5}$ mass concentration.

\subsection{Hazy image synthesis model}
A large-scale hazy image dataset is required for training deep neural networks, but it is not available currently. Nevertheless, large-scale clear image dataset, e.g., the KITTI dataset, is popular and publicly available. The KITTI dataset has been used to validate a bunch of computer vision algorithms and autonomous driving models. Multi-source data have been collected in real-world scenarios, including images, point clouds, and global positioning systems (GPS). The image data were collected by the onboard cameras. This meets the requirement of the presented work. Thus, the KITTI dataset was chosen to synthesize the hazy images, and thus train the proposed framework. \textbf{Proposition 1} is proposed for generating hazy images as follows.

\begin{proposition}
Given an image without haze, $I^{'}\in {\mathbb{R}}^{H\times W \times 3}$, the corresponding depth map, $D\in {\mathbb{R}}^{H\times W \times 1}$, visibility, $V\in {\mathbb{R}}^{1}$, airlight, $A\in {\mathbb{R}}^{1\times 1 \times 3}$, and the minimal observable contrast, $\epsilon$, the hazy image with respect to the specified conditions, $I\in {\mathbb{R}}^{H\times W \times 3}$, can be obtained by the following
\begin{equation}
I=I^{'}\odot e^{\frac{\ln{\epsilon}}{V}R}+A \odot (1-e^{\frac{\ln{\epsilon}}{V}R}),
\label{eq1}
\end{equation}
\begin{equation}
R^{i,j}={\Vert P^{i,j} \Vert}_2,  R\in {\mathbb{R}}^{H\times W \times 1},
\label{eq2}
\end{equation}
\begin{equation}
P^{i,j}=D^{i,j}K^{-1}p,
\label{eq3}
\end{equation}
where $R\in {\mathbb{R}}^{H\times W \times 1}$, $R^{i,j}$, $\odot$, $P\in {\mathbb{R}}^{H\times W \times 3}$, $P^{i,j}$, $K^{-1}\in{\mathbb{R}}^{3\times 3}$, and $p\in {\mathbb{R}}^{3\times 1}$ represent the range image of the scene, the range value at row $i$ and column $j$, element-wise multiplication, the point clouds of the scene, the point cloud at row $i$ and column $j$, the inverse of camera intrinsic matrix, and the pixel coordinate. The broadcasting technique will be used in the course of computation for Eq. \ref{eq1}.
\end{proposition}

\noindent \textbf{Proof.} Light intensity will suffer an attenuation along with the increase of travel distance, because of a series of physical effects, e.g., reflection and scattering. This causes that the observed luminance of objects that are located at various places can be different, even if the actual light intensities of them are identical. With Koschmieder’s Law, the effect can be mathematically stated as 
\begin{equation}
I=I^{'}\odot T+A \odot (1-T),
\label{eq4}
\end{equation}
wherein $T\in {\mathbb{R}}^{H\times W \times 1}$ represents the transmission map of the scene, and is defined as 
\begin{equation}
T=e^{-\beta R}.
\label{eq5}
\end{equation}
In Eq. \ref{eq5}, $e$ and $\beta \in R^1$ represent the natural base and extinction coefficient. In Eq. \ref{eq4}, the airlight, $A$, is defined as the color of light which has been scattered or diffused in the air by dust, haze, or fog (haze is considered in this paper). This color, in general, is regarded as the location-free variable, i.e., the haze colors over different locations in a scene are homogeneous. With priors to the color of haze, it is convenient to generate various $A$ which are close to five possible haze colors, white, blue grey, yellow, grey, and sepia. The key is to determine the transmission map, $T$, which depends on $\beta$ and $R$.

Using any depth estimation model, e.g., Monodepth2 \citep{godard2019digging}, the depth map can then be estimated. Given the depth map of the scene, we can easily get the range map, $R$, by computing the $L_2$ norm for each space coordinate in the point cloud matrix, $P$, which can be obtained by reprojection described in Eq. \ref{eq3}. Therefore, the problem comes to find the extinction coefficient, $\beta$.

As the contrast, $C$, is defined as
\begin{equation}
C=\frac{\hat{p_o}-\hat{p_b}}{\hat{p_b}},
\label{eq6}
\end{equation}
wherein $\hat{p_{o}}$ and $\hat{p_{b}}$ represent the light intensity of object and the baseline intensity. Generally, the horizon sky is chosen for $\hat{p_b}$, i.e., $\hat{p_b}=A$, the contrasts for the clear and hazy images,  
$C_{I^{'}}$ and $C_{I}$, is written as 
\begin{equation}
C_{I^{'}}=\frac{I^{'}-A}{A},
\label{eq7}
\end{equation}
\begin{equation}
\begin{aligned}
C_{I}=\frac{I-A}{A}&=\frac{I^{'}\odot T+A\odot(1-T)-A}{A} \\
&=\frac{I^{'}-A}{A}\odot T=C_{I^{'}}\odot T,
\end{aligned}
\label{eq8}
\end{equation}
where pixel values are directly used as the measurement of light intensity for simplicity.

Given that the minimal observable contrast, $\epsilon$, is defined as the absolute contrast between the black object and airlight, it follows
\begin{equation}
| C_{I^{'}} |=| \frac{0-A}{A}|=1,
\label{eq9}
\end{equation}
\begin{equation}
| C_{I} |=| C_{I^{'}} \odot T|=|-T| \geq \epsilon.
\label{eq10}
\end{equation}
Substituting Eq. \ref{eq5} into Eq. \ref{eq10} furnishes
\begin{equation}
e^{-\beta R} \geq \epsilon.
\label{eq11}
\end{equation}
When equality holds in Eq. \ref{eq11}, the range becomes visibility, i.e.,
\begin{equation}
e^{-\beta R} = \epsilon.
\label{eq12}
\end{equation}
Thus, $\beta$ can be derived accordingly,
\begin{equation}
\beta = - \frac{\ln \epsilon}{V}.
\label{eq13}
\end{equation}
Substituting Eqs. \ref{eq13} and \ref{eq5} into Eq. \ref{eq4} offers Eq. \ref{eq1}.
\begin{flushright}
\qed
\end{flushright}

\begin{remark}
    Although \textbf{Proposition 1} is used to generate haze images using clear images, it also can be used in mimicking other types of contaminated images by simply changing the value of airlight, such as foggy images, snowy images, night images, and so on.
\end{remark}

With \textbf{Proposition 1}, different hazy images, as shown in Figure \ref{kittidepth}, can be gracefully generated based on the images without haze, which will be further used to train the neural networks in the proposed framework.

\subsection{Self-supervised multi-task learning framework}

This subsection introduces a novel bidirectional process of image synthesis and decomposition (BPISD) training pipeline, and thus a self-supervised multi-task learning framework for driving environment perception. Using \textbf{Proposition 1}, a hazy image can be synthesized using the given clear image, depth map, airlight, and visibility. Inspired by this image synthesis process, the opposite image decomposition course can be deployed to estimate those components used for image synthesis. Thus, the following corollary is proposed.

\begin{corollary}
Given a hazy image, $I \in {\mathbb{R}}^{H\times W \times 3}$, and its corresponding clear image, $I^{'} \in {\mathbb{R}}^{H\times W \times 3}$, the range image, $R \in {\mathbb{R}}^{H\times W \times 1}$, airlight, $A \in {\mathbb{R}}^{1 \times 1 \times 3}$, and visibility,  $V \in {\mathbb{R}}^{1}$, of the given hazy image can be estimated by solving the following minimization problem:
\begin{equation}
\begin{aligned}
    &min \quad \Phi (I_r,I) \\
    s.t. \quad &I_r=I^{'} \odot e^{\frac{\ln{\epsilon}}{V}R}+A \odot (1-e^{\frac{\ln{\epsilon}}{V}R})
\end{aligned}
\label{eq14},
\end{equation}
where $\Phi(\cdot)$ represents the function that is to evaluate the difference between the reconstructed image $I_r$ and the input image $I$.
\end{corollary}

\noindent \textbf{Proof.} Based on \textbf{Proposition 1}, the target quantities, $R$, $A$, and $V$, and the given clear image, $I^{'}$, can synthesize haze images. Given a specific set of values $\{R, A, V\}$, the corresponding haze image is determined. By updating $\{R, A, V\}$ towards the direction of minimizing the distance between the synthesized image $I_{r}$ and the original haze image $I$, the resulting solution, $\{R^{*}, A^{*}, V^{*}\}$, will converge to the range image, airlight, and visibility of the given haze image. Therefore, the range image, airlight, and visibility of the given haze image can be estimated.  \begin{flushright} \qed \end{flushright}

\begin{remark}
The problem stated in \textbf{Corollary 1} is a typically ill-posed problem. If $I^{'}$ is not given, which thus has to be estimated, the solution of the presented minimization can collapse to some unexpected cases, e.g., $I^{'}=I$, $R=\textbf{0}$, $A=\textbf{0}$, and $V$ can be any non-zero number. To address this issue, some constraints can be imposed on the estimated either $I^{'}$, $R$, $A$, or $V$. $I^{'}$ was considered known under this circumstance for successfully estimating other quantities, because (1) the corresponding clear image of $I$ can be readily obtained by retrieving historical image data of the same location captured by the on-board cameras outfitted on the CVs in a city; and (2) dehazing has bee well studied, and a number of supervised and unsupervised methods \citep{he2010single, engin2018cycle, yang2018proximal} have been proposed, which can be borrowed to estimate the clear images.
\end{remark}

\begin{remark}
As the range image has the same resolution as compared with the input image and thus is high-dimensional, it is challenging to solve Eq. \ref{eq14} through typical algorithms in the field of optimization. Nevertheless, it is apparent that $R$, $A$, and $V$ are dependent on $I$, meaning that $R$, $A$, and $V$ can somehow be derived from $I$. Despite the impossibility of deriving an explicit relationship between $I$ and $R$/$A$/$V$ (at least it is impossible at this stage), powerful DNNs can be deployed to fit such correlations. Then, it is convenient to train DNNs using gradient-based optimizers.
\end{remark}

Based on \textbf{Corollary 1}, the proposed framework is constituted by multiple tasks, including estimations of range image, airlight, and visibility. Various deep neural networks were used to estimate these quantities. Consider the range estimation network as $\Psi_R$, which is defined as $\Psi_R : I \in {\mathbb{R}}^{H\times W \times 3} \rightarrow V\in {\mathbb{R}}^{1}$. A popular encoder-decoder architecture presented in Monodepth2 \citep{godard2019digging} was adopted for a fair comparison. Differently, single-scale output was utilized instead of four-scale output in the original paper. Similarly, the airlight and visibility estimation networks are denoted as $\Psi_A : I \in {\mathbb{R}}^{H\times W \times 3} \rightarrow A\in {\mathbb{R}}^{1 \times 1 \times 3}$ and $\Psi_V : I \in {\mathbb{R}}^{H\times W \times 3} \rightarrow V\in {\mathbb{R}}^{1}$, respectively. $\Psi_A$ and $\Psi_V$ can either be two separate networks or share the same network with two separate estimation heads. The latter strategy was adopted here. Specifically, ResNet-18 \citep{he2016deep} and four convolutional layers were respectively taken as the encoder and decoder in this paper. To obtain a clear image, the depth estimation network with minor changes to the last layer was used to dehaze in a supervised manner. Consider the dehazing network as $\Psi_C : I \in {\mathbb{R}}^{H\times W \times 3} \rightarrow I^{'}\in {\mathbb{R}}^{H\times W \times 3}$. The training is to solve the following minimization problem:
\begin{equation}
\begin{aligned}
    &min \quad \Theta (I^{'},I^{c}) \\
    s.t. \quad &\Theta (I^{'},I^{c})=\alpha \Vert 
 I^{'}-I^{c} \Vert_{1} + (1-\alpha)\frac{1-SSIM(I^{'},I^{c})}{2},
\end{aligned}
\label{eq15}
\end{equation}
where $SSIM(\cdot)$ and $I^{c}$ represent the structure similarity function and the ground truth clear image; $\alpha$ is set to 0.15 following \citep{mypaper2}. It is noted that the trained dehazing model will be directly used when performing multi-task learning and no longer be trained. As mentioned, clear images can also be estimated by other means, even they can be directly obtained by searching in the historical data.

For each forward process, $\Phi_R$, $\Phi_A$, and $\Phi_V$ take a single hazy image as input to estimate its range image, airlight, and visibility. The trained $\Phi_C$ also takes the hazy image as input and outputs the clear image for further actions. Then, \textbf{Proportion 1} is used to reconstruct the input hazy images based on the estimated $R$, $A$, $I^{'}$ and $V$, as shown in Eq. \ref{eq14}. Let $R_{1}$ and $R_{2}$ be the estimated range images for $I$ and $I^{'}$. Correspondingly, denote the reconstructed images based on $R_1$ and $R_2$ as $I_{r,1}$ and $I_{r,2}$. The reconstruction error can be written as
\begin{equation}
    L_1=\sum\limits_{i=1}^{2} \Theta (I_{r,i},I).
    \label{eq16}
\end{equation}
The estimated ranges images $R_1$ and $R_2$ are expected to be consistent. Thus, the consistency loss is given by 
\begin{equation}
    L_2= \Theta (R_1,R_2).
    \label{eq17}
\end{equation}
Finally, the edge-aware smoothness loss, as follows, is also adopted.
\begin{equation}
    L_3= \beta \sum\limits_{i=1}^{2} | \partial_x \widehat{R}_i^{*}  |e^{- | \partial_x I^{'}|}+ | \partial_y \widehat{R}_i^{*}  |e^{- | \partial_y I^{'}|},
    \label{eq18}
\end{equation}
where $\widehat{R}_i^{*}=\frac{\widehat{R}_i}{E(\widehat{R}_i)}$ is the mean-normalized inverse range following \citep{godard2019digging}; $E(\cdot)$ and $\partial$ represent the expectation operation and the partial derivative operator, respectively; and $\beta$ is set to $0.001$. The final loss is given by 
\begin{equation}
    L= \Phi (I_r,I)=\sum\limits_{i=1}^{3} L_i.
    \label{eq19}
\end{equation}
The Adam optimizer \citep{kingma2014adam} with the initial learning rate of $0.0001$, batch size of $12$, and other default parameters was used to train the networks. The numbers of epochs for dehazing model and multi-task learning framework were set to $100$ and $20$, during which the initial learning rates will be decreased by a factor of $10$ at the $95^{th}$ and the $15^{th}$ epochs. The pretrained model on ImageNet was loaded to initialize the parameters of ResNet-18.

\subsection{Polynomial correlation model for PM$_{2.5}$ mass concentration estimation}
 Applying the trained visibility estimation model, the visibility of a given image can be obtained. To further estimate the PM$_{2.5}$ mass concentration, a simple polynomial correlation model was adopted to model the relationship between visibility and PM$_{2.5}$ mass concentration. It follows that
 \begin{equation}
    \hat{\rho}=\sum_{i=0}^{k}c_{i}V^{i},
    \label{eq21}
\end{equation}
where $\hat{\rho}$, $c_{i}$, and $k$ represent the estimated PM$_{2.5}$  mass concentration, the coefficients of monomials, and the order of the polynomial, respectively. Thus, the key is to estimate the coefficients, $c_{i}, \forall i \in [1, k]$. The problem can be formulated as
 \begin{equation}
    min \sum_{j=1}^{N} (\sum_{i=0}^{k}c_{i}V_{j}^{i}-\rho_{j})^2,
    \label{eq20}
\end{equation}
where $N$ and $\rho_{j}$ represent the number of samples and truthful PM$_{2.5}$  mass concentration. The least square method was utilized to solve the above minimization. The fitted model can then be used to estimate PM$_{2.5}$  mass concentration.

\begin{remark}
    Noting that a small set of truthful PM$_{2.5}$ mass concentrations is available from the sparse meteorological stations, which can be used to calibrate the polynomial correlation model.
\end{remark}

\section{Experiment}

In this section, datasets, definitions of metrics, and experimental results are introduced, respectively. All experiments are implemented with PyTorch $1.9.1$ on a single TITAN RTX card.

\begin{table*}[h]
\caption{Quantitative comparisons of depth estimation on the KITTI dataset \cite{geiger2013vision}. GT and PT represent ground truth and pretraining, respectively. - represents that the situation is unclear. Res. represents resolution. The best performances are marked \textbf{bold}. * The results are reproduced by \citet{mypaper2}. SSD-A, B, C, and D represent different training conditions for depth estimation. A: training depth estimation network with truthful clear images, airlight, and visibility. B: training depth estimation network with truthful clear images and airlight, but estimated visibility. C: training depth estimation network with truthful clear images, but estimated airlight and visibility. D: training depth estimation network with estimated clear image, airlight, and visibility.}
\small
\centering
\begin{tabular}{@{}c|lccccccccc@{}}
\toprule
{\color[HTML]{000000} } & {\color[HTML]{000000} } & {\color[HTML]{000000} } & {\color[HTML]{000000} } & \multicolumn{4}{c}{{\color[HTML]{000000} Errors $\downarrow$}} & \multicolumn{3}{c}{{\color[HTML]{000000} Errors $\uparrow$}} \\ \cline{5-11} 
\multirow{-2}{*}{{\color[HTML]{000000} Res. }} & \multirow{-2}{*}{{\color[HTML]{000000} Methods}} & \multirow{-2}{*}{{\color[HTML]{000000} GT?}} & \multirow{-2}{*}{{\color[HTML]{000000} PT?}} & {\color[HTML]{000000} AbsRel} & {\color[HTML]{000000} SqRel} & {\color[HTML]{000000} RMS} & {\color[HTML]{000000} RMSlog} & {\color[HTML]{000000} $\delta_{1.25}$} & {\color[HTML]{000000} $\delta_{1.25^{2}}$} & {\color[HTML]{000000} $\delta_{1.25^{3}}$} \\ \hline 
{\color[HTML]{000000} } & {\color[HTML]{000000} \citet{eigen2014depth}, coarse} & \multicolumn{1}{c}{{\color[HTML]{000000} $\surd$}} & {\color[HTML]{000000} -}  & {\color[HTML]{000000} 0.214} & {\color[HTML]{000000} 1.605} & {\color[HTML]{000000} 6.563} & {\color[HTML]{000000} 0.292} & {\color[HTML]{000000} 0.673} & {\color[HTML]{000000} 0.884} & {\color[HTML]{000000} 0.957} \\
{\color[HTML]{000000} } & {\color[HTML]{000000} \citet{eigen2014depth}, fine} & \multicolumn{1}{c}{{\color[HTML]{000000} $\surd$}}  & {\color[HTML]{000000} -} & {\color[HTML]{000000} 0.203} & {\color[HTML]{000000} 1.548} & {\color[HTML]{000000} 6.307} & {\color[HTML]{000000} 0.282} & {\color[HTML]{000000} 0.702} & {\color[HTML]{000000} 0.890} & {\color[HTML]{000000} 0.958} \\
{\color[HTML]{000000} } & {\color[HTML]{000000} \citet{liu2015learning}} & \multicolumn{1}{c}{{\color[HTML]{000000} $\surd$}} & {\color[HTML]{000000} -} & {\color[HTML]{000000} 0.202} & {\color[HTML]{000000} 1.614} & {\color[HTML]{000000} 6.523} & {\color[HTML]{000000} 0.275} & {\color[HTML]{000000} 0.678} & {\color[HTML]{000000} 0.895} & {\color[HTML]{000000} 0.965} \\
{\color[HTML]{000000} } & {\color[HTML]{000000} \citet{kuznietsov2017semi}} & \multicolumn{1}{c}{{\color[HTML]{000000} $\surd$}} & {\color[HTML]{000000} $\surd$} & {\color[HTML]{000000} 0.113} & {\color[HTML]{000000} 0.741} & {\color[HTML]{000000} 4.621} & {\color[HTML]{000000} 0.189} & {\color[HTML]{000000} 0.862} & {\color[HTML]{000000} 0.960} & {\color[HTML]{000000} 0.986} \\
\multirow{-5}{*}{{\color[HTML]{000000} -}} & {\color[HTML]{000000} \citet{fu2018deep}} & \multicolumn{1}{c}{{\color[HTML]{000000} $\surd$}} & {\color[HTML]{000000} $\surd$} & {\color[HTML]{000000} \textbf{0.072}} & {\color[HTML]{000000} \textbf{0.307}} & {\color[HTML]{000000} \textbf{2.727}} & {\color[HTML]{000000} \textbf{0.120}} & {\color[HTML]{000000} \textbf{0.932}} & {\color[HTML]{000000} \textbf{0.984}} & {\color[HTML]{000000} \textbf{0.994}} \\ \hline 
{\color[HTML]{000000} } & {\color[HTML]{000000} \citet{yin2018geonet} (VGG)} & {\color[HTML]{000000} } & {\color[HTML]{000000} -} & {\color[HTML]{000000} 0.164} & {\color[HTML]{000000} 1.303} & {\color[HTML]{000000} 6.090} & {\color[HTML]{000000} 0.247} & {\color[HTML]{000000} 0.765} & {\color[HTML]{000000} 0.919} & {\color[HTML]{000000} 0.968} \\
{\color[HTML]{000000} } & {\color[HTML]{000000} \citet{yin2018geonet} (ResNet)} & {\color[HTML]{000000} } & {\color[HTML]{000000} -} & {\color[HTML]{000000} 0.155} & {\color[HTML]{000000} 1.296} & {\color[HTML]{000000} 5.857} & {\color[HTML]{000000} 0.233} & {\color[HTML]{000000} 0.793} & {\color[HTML]{000000} 0.931} & {\color[HTML]{000000} 0.973} \\
{\color[HTML]{000000} } & {\color[HTML]{000000} \citet{wang2018learning}} & {\color[HTML]{000000} } & {\color[HTML]{000000} $\surd$} & {\color[HTML]{000000} 0.151} & {\color[HTML]{000000} 1.257} & {\color[HTML]{000000} 5.583} & {\color[HTML]{000000} 0.228} & {\color[HTML]{000000} 0.810} & {\color[HTML]{000000} 0.936} & {\color[HTML]{000000} 0.974} \\
{\color[HTML]{000000} } & {\color[HTML]{000000} \citet{bian2019unsupervised}} & {\color[HTML]{000000} } & {\color[HTML]{000000} $\surd$} & {\color[HTML]{000000} 0.149} & {\color[HTML]{000000} 1.137} & {\color[HTML]{000000} 5.771} & {\color[HTML]{000000} 0.230} & {\color[HTML]{000000} 0.799} & {\color[HTML]{000000} 0.932} & {\color[HTML]{000000} 0.973} \\
{\color[HTML]{000000} } & {\color[HTML]{000000} \citet{casser2019depth}} & {\color[HTML]{000000} } & {\color[HTML]{000000} -} &  {\color[HTML]{000000} 0.141} & {\color[HTML]{000000} 1.026} & {\color[HTML]{000000} 5.291} & {\color[HTML]{000000} 0.215} & {\color[HTML]{000000} 0.816} & {\color[HTML]{000000} 0.945} & {\color[HTML]{000000} 0.979} \\
{\color[HTML]{000000} } & {\color[HTML]{000000} \citet{mypaper2021}}& {\color[HTML]{000000} } & {\color[HTML]{000000} $\surd$} & {\color[HTML]{000000} 0.144} & {\color[HTML]{000000} \textbf{0.966}} & {\color[HTML]{000000} 5.078} & {\color[HTML]{000000} 0.208} & {\color[HTML]{000000} 0.815} & {\color[HTML]{000000} 0.945} & {\color[HTML]{000000} \textbf{0.981}} \\
{\color[HTML]{000000} } & {\color[HTML]{000000} \citet{klingner2020self}} & {\color[HTML]{000000} } & {\color[HTML]{000000} $\surd$} &  {\color[HTML]{000000} \textbf{0.128}} & {\color[HTML]{000000} 1.003} & {\color[HTML]{000000} 5.085} & {\color[HTML]{000000} 0.206} & {\color[HTML]{000000} 0.853} & {\color[HTML]{000000} 0.951} & {\color[HTML]{000000} 0.978} \\
{\color[HTML]{000000} } & {\color[HTML]{000000} \cite{godard2019digging}} & {\color[HTML]{000000} } & {\color[HTML]{000000} $\surd$} &  {\color[HTML]{000000} \textbf{0.128}} & {\color[HTML]{000000} 1.087} & {\color[HTML]{000000} 5.171} & {\color[HTML]{000000} 0.204} & {\color[HTML]{000000} \textbf{0.855}} & {\color[HTML]{000000} 0.953} & {\color[HTML]{000000} 0.978} \\
{\color[HTML]{000000} } & {\color[HTML]{000000} \citet{mypaper2} (CC)}& {\color[HTML]{000000} } &{\color[HTML]{000000} $\surd$} & {\color[HTML]{000000} \textbf{0.128}} & {\color[HTML]{000000} 0.990} & {\color[HTML]{000000} 5.064} & {\color[HTML]{000000} \textbf{0.202}} & {\color[HTML]{000000} 0.851} & {\color[HTML]{000000} \textbf{0.955}} & {\color[HTML]{000000} 0.980} \\ 
{\color[HTML]{000000} } & {\color[HTML]{000000} \citet{mypaper2} (CL)}& {\color[HTML]{000000} } &{\color[HTML]{000000} $\surd$} & {\color[HTML]{000000} \textbf{0.128}} & {\color[HTML]{000000} 0.979} & {\color[HTML]{000000} \textbf{5.033}} & {\color[HTML]{000000} \textbf{0.202}} & {\color[HTML]{000000} 0.851} & {\color[HTML]{000000} 0.954} & {\color[HTML]{000000} 0.980} \\ \cline{2-11} 
{\color[HTML]{000000} } & {\color[HTML]{000000} \citet{zhou2017unsupervised}} & {\color[HTML]{000000} } & {\color[HTML]{000000} }  & {\color[HTML]{000000} 0.208} & {\color[HTML]{000000} 1.768} & {\color[HTML]{000000} 6.856} & {\color[HTML]{000000} 0.283} & {\color[HTML]{000000} 0.678} & {\color[HTML]{000000} 0.885} & {\color[HTML]{000000} 0.957} \\
{\color[HTML]{000000} } & {\color[HTML]{000000} \citet{yang2017unsupervised}} & {\color[HTML]{000000} } & {\color[HTML]{000000} } &  {\color[HTML]{000000} 0.182} & {\color[HTML]{000000} 1.481} & {\color[HTML]{000000} 6.501} & {\color[HTML]{000000} 0.267} & {\color[HTML]{000000} 0.725} & {\color[HTML]{000000} 0.906} & {\color[HTML]{000000} 0.963} \\
{\color[HTML]{000000} } & {\color[HTML]{000000} \citet{godard2019digging}*} & {\color[HTML]{000000} } & {\color[HTML]{000000} }  & {\color[HTML]{000000} 0.144} & {\color[HTML]{000000} 1.059} & {\color[HTML]{000000} 5.289} & {\color[HTML]{000000} 0.217} & {\color[HTML]{000000} 0.824} & {\color[HTML]{000000} 0.945} & {\color[HTML]{000000} 0.976} \\
{\color[HTML]{000000} } & {\color[HTML]{000000} \citet{mypaper2} (LL)} & {\color[HTML]{000000} } & {\color[HTML]{000000} } & {\color[HTML]{000000} 0.141} & {\color[HTML]{000000} 1.060} & {\color[HTML]{000000} 5.247} & {\color[HTML]{000000} 0.215} & {\color[HTML]{000000} 0.830} & {\color[HTML]{000000} 0.944} & {\color[HTML]{000000} 0.977} \\
{\color[HTML]{000000} } & {\color[HTML]{000000} \citet{jia2023joint}-S} & {\color[HTML]{000000} } & {\color[HTML]{000000} } & {\color[HTML]{000000} 0.135} & {\color[HTML]{000000} 0.973} & {\color[HTML]{000000} 5.084} & {\color[HTML]{000000} 0.208} & {\color[HTML]{000000} 0.840} & {\color[HTML]{000000} 0.948} & {\color[HTML]{000000} 0.978} \\
\multirow{-16}{*}{{\color[HTML]{000000} \rotatebox{90}{$416 \times 128$}}} & {\color[HTML]{000000} \citet{jia2023joint}-L} & {\color[HTML]{000000} } & {\color[HTML]{000000} } & {\color[HTML]{000000} \textbf{0.128}} & {\color[HTML]{000000} \textbf{0.897}} & {\color[HTML]{000000} \textbf{4.905}} & {\color[HTML]{000000} \textbf{0.200}} & {\color[HTML]{000000} \textbf{0.852}} & {\color[HTML]{000000} \textbf{0.953}} & {\color[HTML]{000000} \textbf{0.980}} \\  \hline

{\color[HTML]{000000} } & {\color[HTML]{000000} \citet{godard2019digging}} & {\color[HTML]{000000} } & {\color[HTML]{000000} } & {\color[HTML]{000000} \textbf{0.132}} & {\color[HTML]{000000} 1.044} & {\color[HTML]{000000} 5.142} & {\color[HTML]{000000} 0.210} & {\color[HTML]{000000} \textbf{0.845}} & {\color[HTML]{000000} 0.948} & {\color[HTML]{000000} 0.977} \\
{\color[HTML]{000000} } & {\color[HTML]{000000} \citet{mypaper2} (LL)} & {\color[HTML]{000000} } & {\color[HTML]{000000} } & {\color[HTML]{000000} 0.135} & {\color[HTML]{000000} \textbf{0.979}} & {\color[HTML]{000000} \textbf{5.078}} & {\color[HTML]{000000} \textbf{0.209}} & {\color[HTML]{000000} 0.841} & {\color[HTML]{000000} \textbf{0.949}} & {\color[HTML]{000000} \textbf{0.978}} \\ \cline{2-11}
{\color[HTML]{000000} } & {\color[HTML]{000000} \citet{klingner2020self}} & {\color[HTML]{000000} } & {\color[HTML]{000000} $\surd$} & {\color[HTML]{000000} 0.117} & {\color[HTML]{000000} 0.907} & {\color[HTML]{000000} 4.844} & {\color[HTML]{000000} 0.196 } & {\color[HTML]{000000} 0.875} & {\color[HTML]{000000} 0.958} & {\color[HTML]{000000} 0.980} \\
{\color[HTML]{000000} } & {\color[HTML]{000000} \citet{godard2019digging}} & {\color[HTML]{000000} } & {\color[HTML]{000000} $\surd$} & {\color[HTML]{000000} \textbf{0.115}} & {\color[HTML]{000000} 0.903} & {\color[HTML]{000000} 4.863} & {\color[HTML]{000000} 0.193} & {\color[HTML]{000000} 0.877} & {\color[HTML]{000000} 0.959} & {\color[HTML]{000000} 0.981} \\
{\color[HTML]{000000} } & {\color[HTML]{000000} \citet{mypaper2} (CL)} & {\color[HTML]{000000} } & {\color[HTML]{000000} $\surd$} & {\color[HTML]{000000} 0.116} & {\color[HTML]{000000} 0.886} & {\color[HTML]{000000} 4.787} & {\color[HTML]{000000} 0.192} & {\color[HTML]{000000} 0.876} & {\color[HTML]{000000} 0.959} & {\color[HTML]{000000} 0.981} \\
{\color[HTML]{000000} } & {\color[HTML]{000000} \citet{mypaper2} (CC)} & {\color[HTML]{000000} } & {\color[HTML]{000000} $\surd$} & {\color[HTML]{000000} 0.116} & {\color[HTML]{000000} 0.842} & {\color[HTML]{000000} 4.708} & {\color[HTML]{000000} 0.190} & {\color[HTML]{000000} 0.876} & {\color[HTML]{000000} 0.961} & {\color[HTML]{000000} 0.982} \\
{\color[HTML]{000000} } & {\color[HTML]{000000} \citet{mypaper2} (CC)} & {\color[HTML]{000000} } & {\color[HTML]{000000} $\surd$} & {\color[HTML]{000000} 0.116} & {\color[HTML]{000000} 0.842} & {\color[HTML]{000000} 4.708} & {\color[HTML]{000000} 0.190} & {\color[HTML]{000000} 0.876} & {\color[HTML]{000000} 0.961} & {\color[HTML]{000000} 0.982} \\
{\color[HTML]{000000} } & {\color[HTML]{000000} Our (SSD-A)} & {\color[HTML]{000000} } & {\color[HTML]{000000} $\surd$} & \textbf{0.115} &	0.866 &	4.643 &	0.188 &	\textbf{0.878} &	0.962 &	 0.983 \\
{\color[HTML]{000000} } & {\color[HTML]{000000} Our (SSD-B)} & {\color[HTML]{000000} } & {\color[HTML]{000000} $\surd$} & 0.116 &	0.836 &	\textbf{4.607} &	\textbf{0.187} &	\textbf{0.878} &	\textbf{0.963} &	0.983 \\
{\color[HTML]{000000} } & {\color[HTML]{000000} Our (SSD-C)} & {\color[HTML]{000000} } & {\color[HTML]{000000} $\surd$} & 0.117 &	0.858 &	4.635 &	0.189 &	0.876 &	\textbf{0.963} &	0.983 \\
\multirow{-9}{*}{{\color[HTML]{000000} \rotatebox{90}{$640 \times 192$}}} & {\color[HTML]{000000} Our (SSD-D)} & {\color[HTML]{000000} } & {\color[HTML]{000000} $\surd$} & 0.122 &	\textbf{0.810} &	4.611 &	0.188 &	0.870 &	\textbf{0.963} &	\textbf{0.984} \\
\bottomrule
\end{tabular}
\label{kittiresults}
\end{table*}

\begin{table*}[h]
\caption{Quantitative results of self-supervised depth estimation under different visibility conditions. SSD-A, B, C, and D represent different training conditions for depth estimation. A: training depth estimation network with truthful clear images, airlight, and visibility. B: training depth estimation network with truthful clear images and airlight, but estimated visibility. C: training depth estimation network with truthful clear images, but estimated airlight and visibility. D: training depth estimation network with estimated clear image, airlight, and visibility.}
\small
\centering
\begin{tabular}{@{}ccccccccc@{}}
\toprule

{\color[HTML]{000000} } & {\color[HTML]{000000} } & \multicolumn{4}{c}{{\color[HTML]{000000} Errors $\downarrow$}} & \multicolumn{3}{c}{{\color[HTML]{000000} Errors $\uparrow$}} \\ \cline{3-9} 
\multirow{-2}{*}{{\color[HTML]{000000} Methods}} & \multirow{-2}{*}{{\color[HTML]{000000} Visibility}} & {\color[HTML]{000000} AbsRel} & {\color[HTML]{000000} SqRel} & {\color[HTML]{000000} RMS} & {\color[HTML]{000000} RMSlog} & {\color[HTML]{000000} $\delta_{1.25}$} & {\color[HTML]{000000} $\delta_{1.25^{2}}$} & {\color[HTML]{000000} $\delta_{1.25^{3}}$} \\ \hline 
\citet{godard2019digging}&	Perfect&	\textbf{0.115}&	0.903&	4.863&	0.193&	0.877&	0.959&	0.981\\ 
SSD-A&	Perfect&	\textbf{0.115}&	0.866&	4.643&	0.188&	\textbf{0.878}&	0.962&	 0.983\\ 
SSD-B&	Perfect&	0.116&	0.836&	\textbf{4.607}&	\textbf{0.187}&	\textbf{0.878}&	\textbf{0.963}&	0.983\\ 
SSD-C&	Perfect&	0.117&	0.858&	4.635&	0.189&	0.876&	\textbf{0.963}&	0.983\\ 
SSD-D&	Perfect&	0.122&	\textbf{0.810}&	4.611&	0.188&	0.870&	\textbf{0.963}&	\textbf{0.984}\\ \hline

\citet{godard2019digging}&	Good&	0.125&	0.994&	5.147&	0.203&	0.860&	0.954&	0.979\\
SSD-A&	Good&	\textbf{0.115}&	0.892&	4.686&	\textbf{0.188}&	\textbf{0.878}&	0.962&	0.983\\
SSD-B&	Good&	0.117&	0.851&	4.653&	\textbf{0.188}&	\textbf{0.878}&	\textbf{0.963}&	0.983\\
SSD-C&	Good&	0.118&	0.873&	4.682&	0.189&	0.875&	0.962&	0.983\\
 SSD-D&	Good&	0.121&	\textbf{0.834}&	\textbf{4.632}&	\textbf{0.188}&	0.870&	\textbf{0.963}&	\textbf{0.984}\\ \hline

\citet{godard2019digging}&	Middle&	0.138&	1.063&	5.526&	0.217&	0.835&	0.945&	0.977\\
SSD-A&	Middle&	\textbf{0.116}&	0.876&	4.683&	\textbf{0.188}&	0.877&	0.962&	0.983\\
SSD-B&	Middle&	0.117&	0.870&	4.688&	\textbf{0.188}&	\textbf{0.878}&	0.962&	0.983\\
SSD-C&	Middle&	0.118&	0.889&	4.720&	0.190&	0.875&	0.962&	0.983\\
SSD-D&	Middle&	0.121&	\textbf{0.852}&	\textbf{4.660}&	0.189&	0.870&	\textbf{0.963}&	\textbf{0.984}\\ \hline

\citet{godard2019digging}&	Bad&	0.161&	1.278&	6.253&	0.244&	0.787&	0.928&	0.968 \\
SSD-A&	Bad&	\textbf{0.116}&	0.887&	4.709&	\textbf{0.189}&	0.876&	0.961&	\textbf{0.983} \\
SSD-B&	Bad&	0.117&	0.886&	4.703&	\textbf{0.189}&	\textbf{0.877}&	\textbf{0.962}&	0.982\\
SSD-C&	Bad&	0.119&	0.886&	4.733&	0.191&	0.874&	0.961&	\textbf{0.983} \\
SSD-D&	Bad&	0.122&	\textbf{0.868}&	\textbf{4.691}&	0.190&	0.869&	\textbf{0.962}&	\textbf{0.983} \\ \hline

\citet{godard2019digging} &	Terrible&	0.270&	2.478&	8.554&	0.358&	0.571&	0.826&	0.915\\
SSD-A &	Terrible&	\textbf{0.117}&	0.926&	4.813&	\textbf{0.191}&	\textbf{0.874}&	\textbf{0.960}&	0.982\\ 
SSD-B&	Terrible&	0.120&	0.907&	\textbf{4.772}&	0.192&	0.873&	\textbf{0.960}&	0.982\\ 
SSD-C&	Terrible&	0.123&	0.960&	4.940&	0.195&	0.868&	0.959&	0.982\\ 
SSD-D&	Terrible&	0.129&	\textbf{0.867}&	4.803&	0.196&	0.854&	0.959&	\textbf{0.983} \\ \hline
 \bottomrule
\end{tabular}
\label{kitti2}
\end{table*}

\begin{figure*}[htp]
    \centering
    \includegraphics[width=\textwidth]{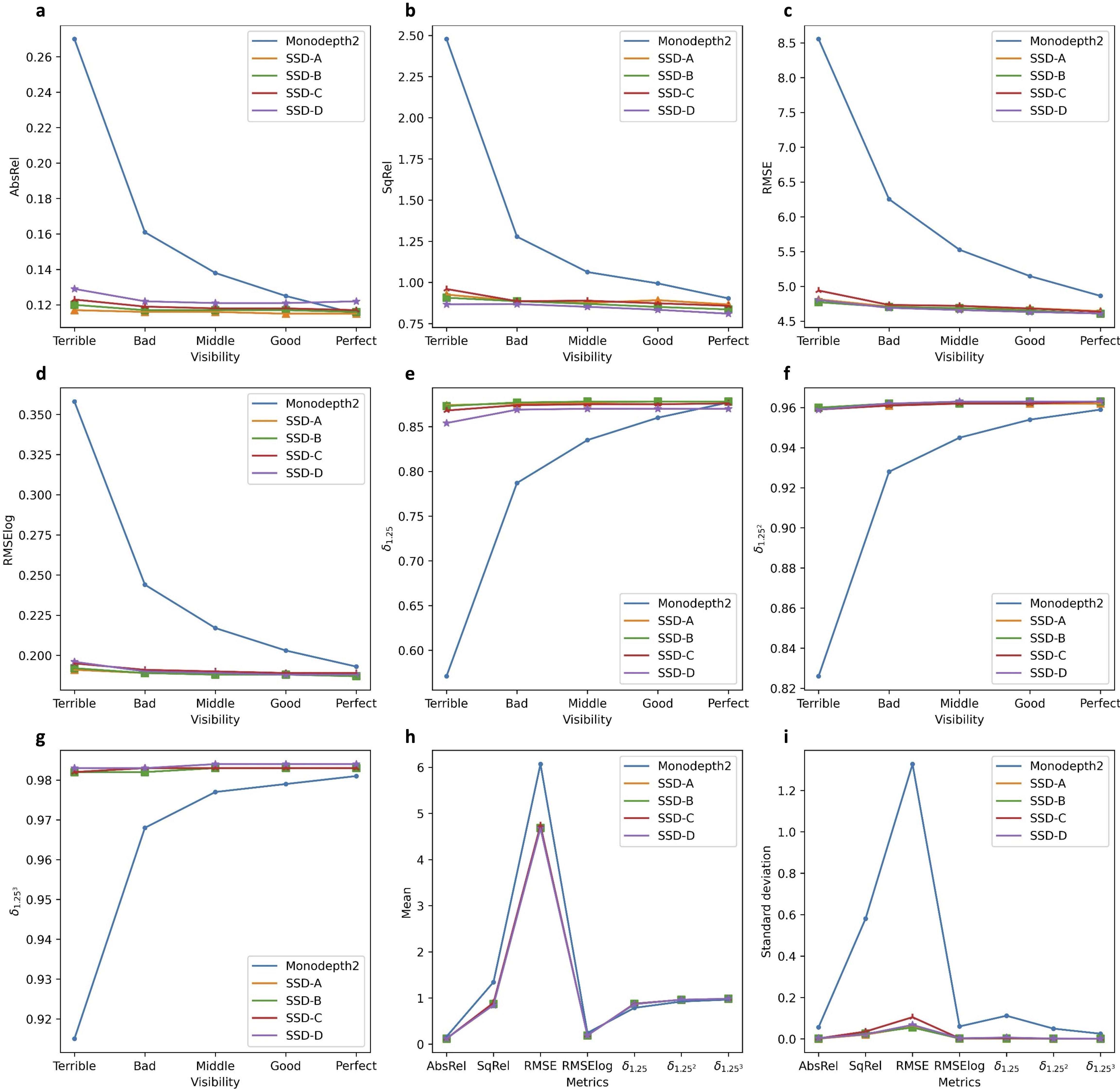}
    \caption{Results of self-supervised depth estimation on the KITTI dataset \citep{geiger2013vision}. The results on various evaluation metrics, including AbsRel, SqRel, RMSE, RMSElog, $\delta_{1.25}$, $\delta_{1.25^{2}}$, $\delta_{1.25^{3}}$, are shown in a, b, c, d, e, f, and g, respectively. h and i represent the means and standard deviations of different metrics across various visibility conditions.}
    \label{kittidepthnumber}
\end{figure*}

\renewcommand\floatpagefraction{.9}
\begin{figure*}[htp]
    \centering
    \includegraphics[width=\textwidth]{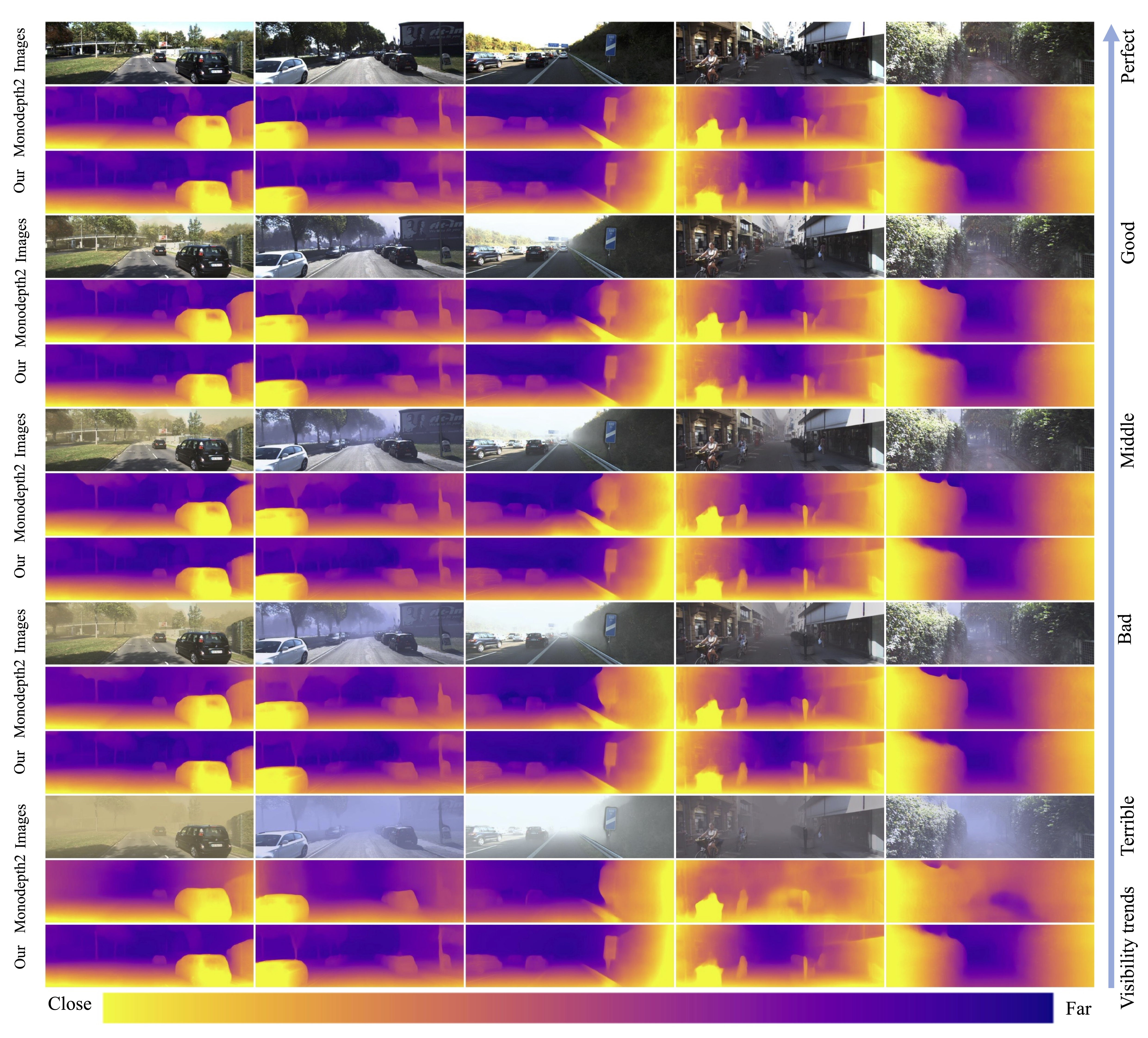}
    \caption{Qualitative comparisons of depth estimation under various visibility conditions on the KITTI dataset \citep{geiger2013vision}.}
    \label{kittidepth}
\end{figure*}

\subsection{Datasets}

For self-supervised depth estimation, a large-scale and popular dataset, KITTI \citep{geiger2013vision}, was used to train the networks. Based on Zhou et. al’ split \citep{zhou2017unsupervised} on the KITTI dataset, $40,109$ images with various visibility conditions were first generated in terms of \textbf{Proposition 1} to mimic the different levels of PM$_{2.5}$ mass concentrations. Five visibility scales, i.e., Terrible, Bad, Middle, Good, and Perfect, were considered with respect to the corresponding relative visibilities of 0.1, 0.3, 0.5, 0.8, and 1, where 1 represents the clear images without haze. In each visibility group, $697$ images were used for testing. The image resolution is set to $640 \times 192$.

It is noted that some images in the original test set of Eigen’s split are not suitable for evaluating visibility and airlight as the scenes are extremely close to the cameras or are full of buildings and pedestrians. Thus, in visibility and airlight estimation, $377$ images in Eigen’s split \citep{zhou2017unsupervised} of the KITTI dataset were chosen for generating testing images, which are synthesized based on the given clear images and randomly generated relative visibilities in the range of $0$ to $1$.

To validate the effectiveness of the proposed method in estimating PM$_{2.5}$ mass concentration, real-world images with various PM$_{2.5}$ mass concentrations were collected. The image data were manually selected on the Beijing Tour website, on which the real-time image data and detailed weather information are provided, including temperature, humidity, wind, etc. The corresponding PM$_{2.5}$ mass concentration of each image is obtained from the U.S. Embassy in Beijing. The numbers of images in relative humidity 0 to 0.5, 0.5 to 0.7, 0.7 to 0.9, and 0.9 to 1 are 39, 17, 13, and 24, respectively, i.e., 93 images with different PM$_{2.5}$ mass concentrations were used for validation. 

\subsection{Definitions of metrics}
This subsection introduces the metrics used for evaluations. Denote the ground truth depth map and the predicted depth map as $D^{gt} \in \mathbb{R}^{h \times w}$ and $D^{pre} \in \mathbb{R}^{h \times w}$, where $h$ and $w$ are the height and width of the depth map, respectively. $N$ is the number of valid pixels in the ground truth depth map. Thus, the metrics used for evaluating depth estimation are defined as follows:
\begin{itemize}
\item Absolute relative (AbsRel) error (Eq. \ref{AbsRel}):
\begin{equation}
  \label{AbsRel}
  AbsRel = \frac{1}{N} \sum_{i=1}^{N} \frac{|D^{pre}_{i} - D^{gt}_{i}|}{D^{gt}_{i}};
\end{equation}
\item Square relative (SqRel) error (Eq. \ref{SqRel}):
\begin{equation}
  \label{SqRel}
  SqRel = \frac{1}{N} \sum_{i=1}^{N} \frac{|D^{pre}_{i} - D^{gt}_{i}|^{2}}{D^{gt}_{i}};
\end{equation}
\item Root mean square (RMS) error (Eq. \ref{RMS}):
\begin{equation}
  \label{RMS}
  RMS = \sqrt{\frac{1}{N} \sum_{i=1}^{N} |D^{pre}_{i} - D^{gt}_{i}|^{2}};
\end{equation}
\item Root mean square logarithm (RMSlog) error (Eq. \ref{RMSlog}):
\begin{equation}
  \label{RMSlog}
  RMSlog = \sqrt{\frac{1}{N} \sum_{i=1}^{N} |log D^{pre}_{i} - log D^{gt}_{i}|^{2}};
\end{equation}
\item $\delta_{T}$ accuracy (Eq. \ref{accuracy}):
\begin{equation}
  \label{accuracy}
  \begin{aligned}
  \delta_{T} &= \frac{\sum_{i=1}^{N} (max(\frac{D^{pre}_{i}}{D^{gt}_{i}}, \frac{D^{gt}_{i}}{D^{pre}_{i}}) < T)}{N}\\
  T &= 1.25, 1.25^{2}, 1.25^{3}
  \end{aligned}.
\end{equation}
\end{itemize}

In the evaluation of visibility, airlight, and  PM$_{2.5}$ mass concentration estimation, RMS error (RMSE), mean absolute error (MAE), and mean absolute percentage error (MAPE) were adopted. For each test item, the metrics stated above are computed. Then, the final results are obtained by averaging all testing data.

\begin{table*}[h]
\caption{Quantitative results of PM$_{2.5}$ mass concentration estimation on the real-world data. RH represents relative humidity. Order represents the polynomial fitting order.}
\small
\centering
\begin{tabular}{@{}ccccccccccccc@{}}
\toprule
 & \multicolumn{3}{c}{$0 \leq RH \textless 0.5$} & \multicolumn{3}{c}{$0.5 \leq RH \textless 0.7$} & \multicolumn{3}{c}{$0.7 \leq RH \textless 0.9$} & \multicolumn{3}{c}{$0.9 \leq RH \textless 1$}  \\ 
\cline{2-13}
\multirow{-2}{*}{{\color[HTML]{000000} Order}} & RMSE&	MAE&	MAPE($\%$)&	RMSE&	MAE&	MAPE($\%$)&	RMSE&	MAE&	MAPE($\%$)&	RMSE&	MAE&	MAPE($\%$) \\
\hline
1&	55.8&	36.8&	16.0&	58.2&	49.7&	13.0&	61.0&	48.7&	13.4&	42.6&	33.3&	5.9\\

2&	37.4&	26.4&	11.5&	47.9&	38.5&	10.3&	43.0&	36.4&	10.6&	37.6&	30.9&	5.4 \\

3&	22.4&	18.5&	8.9&	43.9&	36.8&	9.7&	18.3&	14.6&	3.9&	37.6&	30.9&	5.4\\

4&	21.2&	17.2&	8.2&	40.9&	34.3&	9.0&	18.1&	14.3&	3.8&	32.9&	25.9&	4.5\\

5&	20.1&	15.1&	7.1&	40.8&	34.2&	9.0&	16.7&	13.9&	4.0&	32.9&	26.0&	4.5\\

6&	19.2&	13.5&	6.3&	40.3&	30.7&	7.9&	12.6&	9.9&	2.9&	30.4&	24.1&	4.1\\

7&	18.9&	13.4&	6.2&	40.1&	31.0&	8.0&	11.6&	7.1&	2.9&	30.3&	23.8&	4.1 \\

8&	18.9&	13.4&	6.2&	35.1&	27.5&	7.3&	11.6&	7.1&	2.1&	30.2&	23.7&	4.1\\

9&	18.8&	13.0&	6.0&	34.2&	27.0&	7.2&	11.6&	7.1&	2.1&	30.1&	23.4&	4.0\\

10&	18.7&	13.3&	6.2&	33.7&	26.9&	7.3&	10.7&	6.1&	1.8&	29.1&	22.6&	3.8\\
 \bottomrule
\end{tabular}
\label{pm2.5number}
\end{table*}

\subsection{Results}
The proposed co-training system is constituted by four sub-tasks: estimations of the depth map, airlight, visibility, and PM2.5 mass concentration. All these four tasks are evaluated.

\subsubsection{Self-supervised depth estimation}

Current self-supervised depth estimation systems are based on view synthesis and use reference images to reconstruct the target image. The reconstruction error is taken as the loss to train the networks. Such a view-synthesis approach requires an image sequence as input for training. However, videos may not be always available, due to limited transmission bandwidth and energy supply on edge devices. Instead, the presented self-supervised depth estimation system solely takes a single image as input, i.e., performing single-image and self-supervised depth (SSD) estimation. This provides a new paradigm for conducting monocular depth estimation. 

Table \ref{kittiresults} compares the proposed method with other state-of-the-art methods on the original KITTI dataset. Apparently, the proposed method achieves performance competitive to other methods on all evaluation metrics. "SSD-A, B, C, and D" show the ablation studies by gradually relaxing the training conditions, which indicate that the proposed system can accurately estimate the depth map as well as the byproducts (i.e., visibility and airlight), even with the estimated clear images.

Table \ref{kitti2} and Figure \ref{kittidepthnumber} reports the quantitative results of SSD under various visibility conditions on the KITTI dataset. Detailed ablation studies on the training conditions were also presented. “SSD–A” model was trained with truthful clear images, airlight, and visibility, which outperforms Monodepth2 \citep{godard2019digging} on various visibility conditions. In particular, low-visibility conditions significantly degrade the performance of Monodepth2. In contrast, the proposed method solely undergoes neglectable performance drops. “SSD–B, C, and D” gradually relax the training condition to without truthful visibility, without truthful visibility and airlight, and finally without truthful visibility, airlight, and clear image. All these unknown data were simultaneously estimated through the proposed framework. The results indicate that the performance only slightly drops as compared with that of “SSD–A”. For a perfect visibility situation, the performance of the final model “SSD–D” is still on par with that of Monodepth2 \citep{godard2019digging}. In other hazy conditions, “SSD–D” exhibits huge advantages on all metrics as compared with Monodepth2 \citep{godard2019digging}. This clearly demonstrates the effectiveness and robustness of the proposed framework.

Some qualitative comparisons across various visibility conditions are presented in Figure \ref{kittidepth}. Given the perfect-visibility images, the qualitative results from the proposed method and Monodepth2 \citep{godard2019digging} are highly consistent. Nevertheless, with the innovative training framework, the proposed method performs much better than Monodepth2 on some difficult regions, e.g., windows. With visibility deteriorating, many details and far scenes could not be estimated in Monodepth2. The proposed method still provides accurate estimations with vivid details.

\begin{figure*}[htp]
    \centering
    \includegraphics[width=\textwidth]{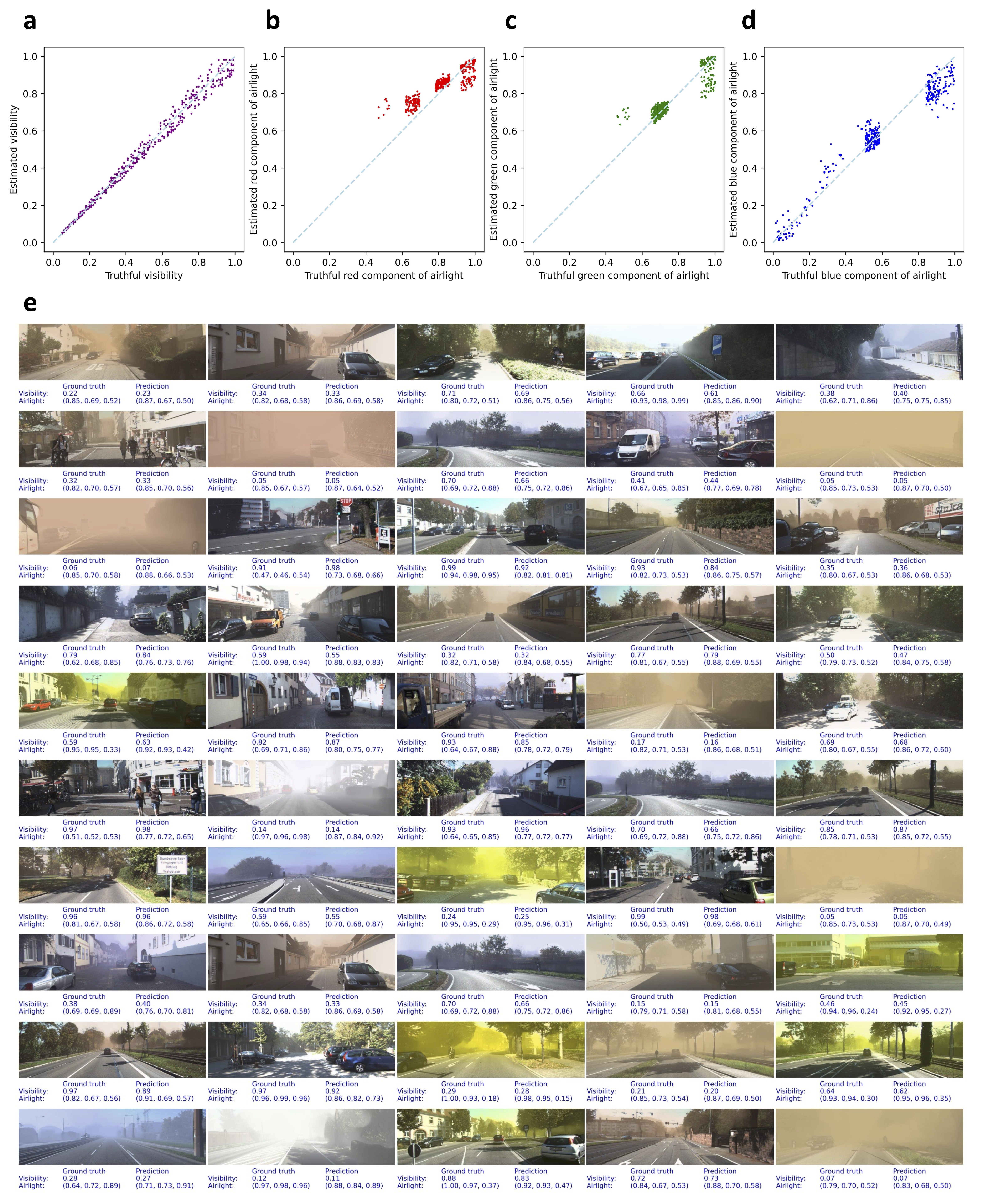}
    \caption{Results of self-supervised visibility and airlight estimations on the KITTI dataset \citep{geiger2013vision}. a: the estimated visibility versus ground truth visibility. b/c/d: the estimated red/green/blue component of airlight versus truthful red/green/blue component. e: some estimation samples.}
    \label{kittivisibility}
\end{figure*}

\begin{figure*}[htp]
    \centering
    \includegraphics[width=\textwidth]{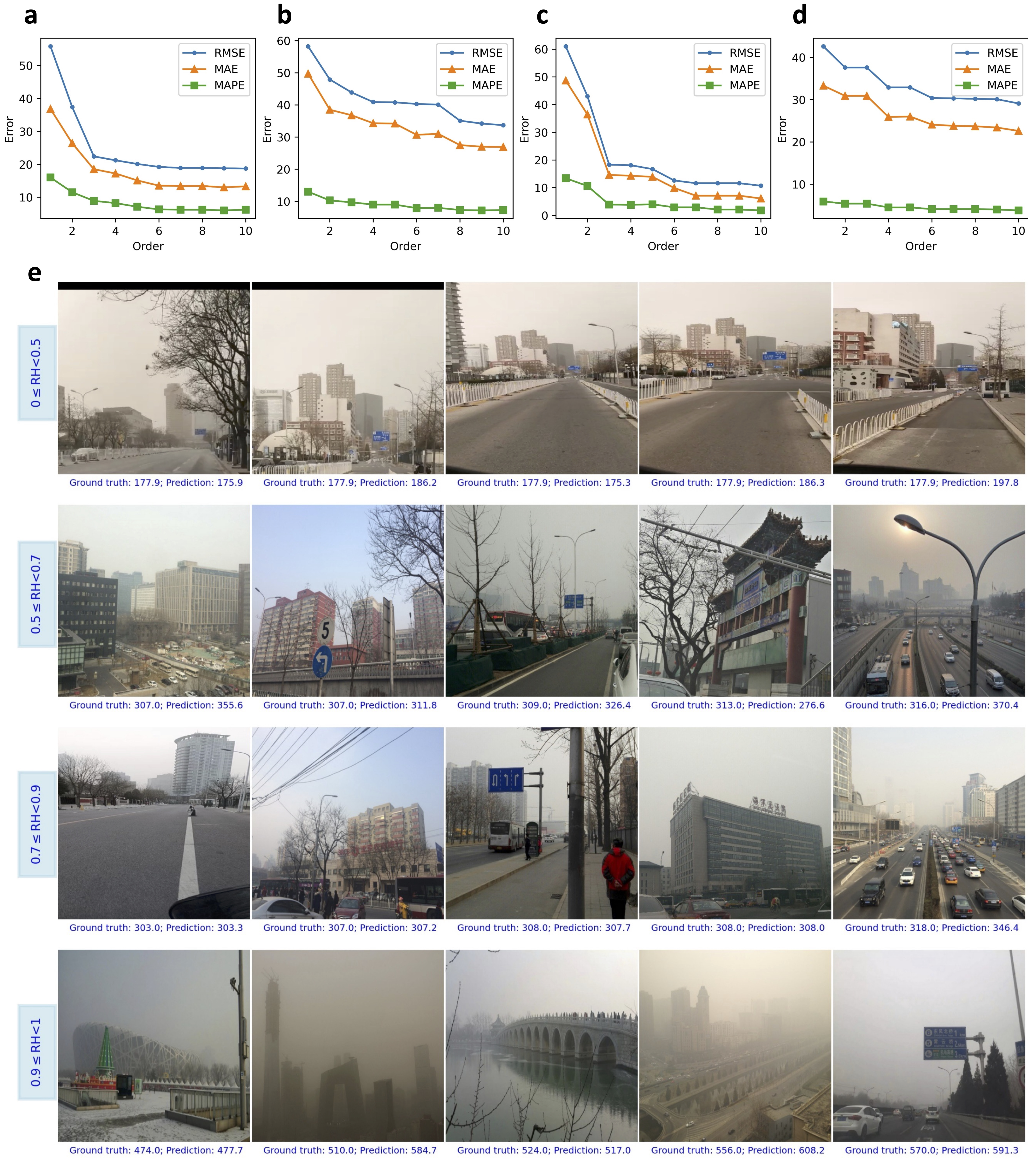}
    \caption{Results of PM$_{2.5}$ mass concentration estimation on the real-world data. a, b, c, d: estimation errors with the relative humidity of 0 to 0.5, 0.5 to 0.7, 0.7 to 0.9, and 0.9 to 1. e: some estimation examples (the unit for numbers is $\mu g/m^{3}$)}
    \label{pm2.5}
\end{figure*}

\subsubsection{Self-supervised visibility and airlight estimation}
Visibility for a given image is represented by a scalar, which is considered highly associated with traffic management and safety. Applying the trained self-supervised visibility estimation model to the selected KITTI data, it was found that the RMSE, MAE, and MAPE of the proposed method are 0.032, 0.025, and $4.9\%$, respectively. 

Airlight, which is represented by a 3D vector (i.e., red, green, and blue components), is caused by light scattering and diffusion by particulate matters, e.g., dust and haze. Airlight owns spatiotemporal variations due to the changes in chemical components in the air. Conversely, knowledge of airlight plays a critical role in analyzing the chemical components in the air and thus provides important guidance to improve air quality and more precise warnings to the public. Applying the trained airlight estimation model to the same set of 337 images used for self-supervised visibility estimation, the RMSE, MAE, and MAPE were found to be 0.067, 0.049, and $9.2\%$, respectively. Figures \ref{kittivisibility}a, \ref{kittivisibility}b, \ref{kittivisibility}c, and \ref{kittivisibility}d show the estimated quantities versus the ground truth quantities graphs. The dots are well distributed along the 45-degree lines, meaning that the estimations are sufficiently accurate. Figure \ref{kittivisibility}e presents some estimation examples. The minor prediction deviations clearly demonstrate the effectiveness of the proposed framework on self-supervised visibility and airlight estimation.

\subsubsection{PM$_{2.5}$ mass concentration estimation}
Assuming that low visibility is solely caused by PM$_{2.5}$, PM$_{2.5}$ mass concentration can be derived from the atmosphere visibility. By directly applying the trained visibility estimation model on the KITTI dataset to real-world data, the relative visibilities were first estimated. Then, the polynomial correlation model was used to match the estimated relative visibilities with the truthful PM$_{2.5}$ mass concentrations.

As shown in Table \ref{pm2.5number}, Figures \ref{pm2.5}a, \ref{pm2.5}b, \ref{pm2.5}c, and \ref{pm2.5}d, the estimation performance gradually gets improved and being asymptotically stable with the increase of polynomial order. The absolute percentage errors were well confined within $8\%$ as the polynomial order is equal to or greater than 6. RMSE and MAE were respectively confined within 41 and 31 over various humidity conditions. Figure \ref{pm2.5}e presents some estimation results by setting the polynomial order to 10. Excellent performance indicates that the proposed method can accurately estimate PM$_{2.5}$ mass concentrations under various humidity conditions. It has great potential to be implemented in an urban system to dynamically monitor PM$_{2.5}$ mass concentrations.

\section{Conclusion}

This paper proposes a novel framework to simultaneously conduct the estimations of the depth map, airlight, visibility, and PM$_{2.5}$ mass concentrations leveraging the CV technologies. The on-road CVs can share local information with other vehicles and data centers. Image data collected by the onboard cameras are used in this study. Due to the different trips of CVs, it is assumed that CVs are well distributed in the city. Consider a specific time instant; the images in different urban regions can be obtained via CVs. Along with the traverse of CVs in the whole city, the local images can be regularly updated. Moreover, the spatial resolution can be readily adjusted by setting various data report frequencies of CVs. Real-time precise airlight, visibility, PM$_{2.5}$ mass concentrations, and depth maps can be estimated accordingly.

The proposed framework solely requires a single input image without any labels for all sub-tasks, working in a self-supervised manner. Moreover, it is a non-intrusive method, meaning that no additional equipment, professional instruments, and special adjustments to vehicles are involved. Thus, it is considered more convenient and flexible as compared with other methods in estimating those quantities. Comprehensive experiments demonstrate the effectiveness and superiority of the proposed method. In depth estimation, the proposed method achieves performance competitive to current self-supervised methods when taking clear images as input, and significantly outperforms current methods when taking hazy images as input. Minor estimation derivations on airlight, visibility, and PM$_{2.5}$ mass concentration further show its great application potential. We expect to develop real-world applications in the future.

\section*{Acknowledgement}
The work described in this paper was partially supported by the National Natural Science Foundation of China (Project No. 42171361), The Hong Kong Polytechnic University under Projects 1-ZVN6, 1-ZECE, and Q-CDAU. 
\bibliography{mybibfile}

\end{document}